# Effect of secular trend in drug effectiveness study in real world data

**Sharon Hensley Alford**[*]    **Piyush Madan**[*]    **Shilpa Mahatma**    **Italo Buleje**
**Yanyan Han    Fang Lu**
IBM Watson                             IBM Research
shenlse@us.ibm.com        {Piyush.Madan1, mahatma, ibuleje, yanyan.han, fang.lu}@us.ibm.com

## Abstract

We discovered secular trend bias in a drug effectiveness study for a recently approved drug. We compared treatment outcomes between patients who received the newly approved drug and patients exposed to the standard treatment. All patients diagnosed after the new drug's approval date were considered. We built a machine learning causal inference model to determine patient subpopulations likely to respond better to the newly approved drug. After identifying the presence of secular trend bias in our data, we attempted to adjust for the bias in two different ways. First, we matched patients on number of days from the new drug's approval date that the patient's treatment (new or standard) began. Second, we included a covariate in the model for number of days between the date of approval of the new drug and the treatment (new or standard) start date. Neither approach completely mitigated the bias. Residual bias we attribute to differences in patient disease severity or other unmeasured patient characteristics. Had we not identified the secular trend bias in our data, the causal inference model would have been interpreted without consideration for this underlying bias. Being aware of, testing for, and handling potential bias in the data is essential to diminish the uncertainty in AI modeling.

_________________

* equal contribution

## 1. INTRODUCTION

Patient medical data is expanding in scope and depth. Advances in mining unstructured text will eventually surface the estimated 80% of patient data currently unavailable in clinical notes. [1] In addition, ~omics data is expected to become readily available. The application of AI approaches to medical data is expected to result in significant advances in medicine. [2,3,4] However, there is distrust of AI in medicine and uncertainty about the results and insights generated with machine learning algorithms. [5] Data scientists working in medicine need to be particularly vigilant about bias and confounding since patient level decisions could eventually be influenced by model results. [6] In this paper, we expose an important bias in the use of prescription fill data. And demonstrate that consideration of drug adoption patterns in medical practice need to be assessed when comparing treatment effects in patient data.

## 2. METHODS

In a drug effectiveness study using real world data, we studied a drug which was recently approved for first line treatment and was the first drug in a new drug class. In our study design, we included patients who either took the newly approved drug (Drug A) or the standard treatment (Drug B). All the patients in the cohort were diagnosed at the time of or after Drug A's approval. Patients were given either Drug A or Drug B. The observation started at the beginning of treatment and ended at either death of the patient, disenrollment from the health plan, or end of the study, which ever came first. Outcome of interest was lack of treatment effectiveness (LTE) as defined by clinical experts.

We tested two approaches to control for the bias: matching patients on how many days from the new drug's approval the patient's treatment began and adding a covariate to the model for the number of days from the new drug's approval and treatment initiation. A third approach, that was not feasible in our data, would be to compare patients only after prescribing behavior suggested full clinical adoption of the new drug. We also assessed the prescribing patterns in a selection of other newly approved drugs to determine if our observation was unique.

## 2.1 MATCHING PATIENTS

Both treatment groups included patients who initiated treatment after the date of approval of Drug A. The ratio of patients taking Drug A versus Drug B changed over time. We matched patients receiving Drug A with patients receiving Drug B on date of treatment initiation ± 30 days using a 1 to 3 ratio (new to standard drug). When more than 3 matches were available we randomly selected the matches.

## 2.2 USING A TEMPORAL FEATURE

We included a model covariate in our causal inference models for the number of days from Drug A's approval and when treatment initiation began. We also built machine learning models and calculated the feature importance for days between drug approval date and date when the patient started treatment.

## 2.3 STARTING THE OBSERVATION PERIOD AFTER THE END OF SECULAR TREND

Another way to overcome secular trend bias is to start the study's observation period after clinical adoption has occurred. In our study, the recency of the drug's approval did not allow a long enough follow-up period to construct causal inference models on treatment effectiveness this way. However, in theory, this approach would control for the adoption effect as well as the potential that treatment assignment is based on disease severity.

## 2.4 ASSESSING SECULAR TRENDS IN OTHER NEWLY APPROVED DRUGS

To determine if our observations of secular trend bias in Drug A were unique, we selected three drugs approved in similar years to Drug A. Using the same dataset, we looked at all prescriptions filled from the drug's approval date. Then we assessed the patterns that emerged for each drug.

## 3. RESULTS

A total of 939 patients were prescribed Drug A and 1833 patients were prescribed Drug B. All patients were the same sex. Race/ethnicity was not available in the data set. The mean number of observation days was μ=366 (sd=230.6) for patients taking Drug A and μ=422 (sd=249.36) for patients taking Drug B. The t-test for difference in means was 2.97, p-value 0.0029.

Figures 1 and 2 represent the number of patients receiving each drug over time with trend lines superimposed.

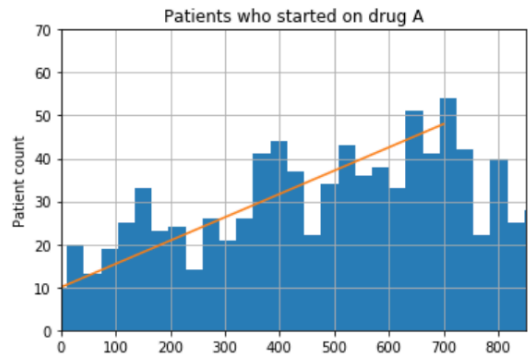

*Figure 1: Distribution of patients who started Drug A. X-axis shows days from the approval of Drug A that treatment was initiated and Y-axis shows number of patients initiating treatment on that day. Orange line shows the predominant pattern of Drug A prescribing.*

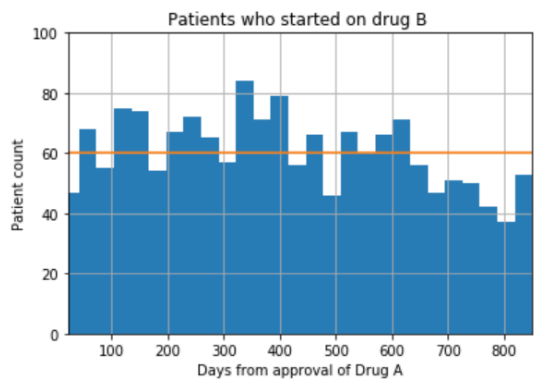

*Figure 2: Distribution of patients who started Drug B. X-axis shows days from Drug A's approval that treatment was initiated and Y-axis shows number of patients initiating treatment on that day. Orange colored line is to show the predominant pattern of Drug B prescribing.*

Figure 3 shows the characteristics that separate responders to Drug A vs. Drug B in our population using the Ozery-Flato method. [7] When separating the population into responders for Drug A compared to responders to Drug B, days from Drug A's approval remains influential in defining patients' treatment response. Those who responded better to Drug A were assigned that treatment well over a year after the drug's approval. While patients who did better on Drug B were more likely to receive the standard treatment in the 6-8 months following the new treatment's availability.

Results of our causal inference model show the continued influence of the bias even time from approval is included in the model as a covariate.

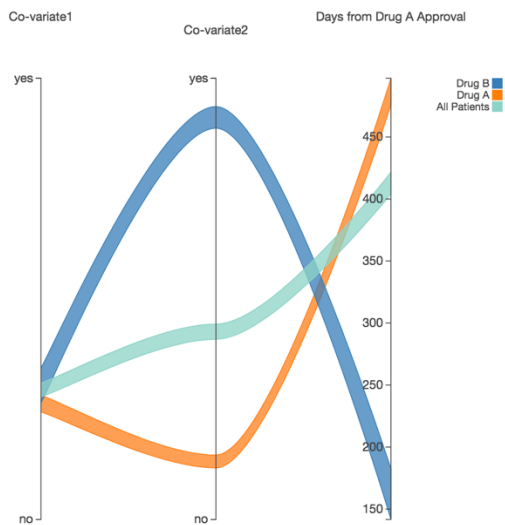

*Figure 3: Influence of effect modifiers in causal inference model*

Figures 4-6 show the prescribing patterns observed in the 24 months following approval for three drugs approved in the same time period as Drug A for various conditions. These include Acyclovir (antiviral, BioAlliance Pharma, April 2013), Dolutegravir (anti-retroviral, ViiV Healthcare, August 2013), and Lorcaserin (weight management, Arena Pharmaceuticals, June 2012). There is considerable variation in clinical adoption of these new drugs.

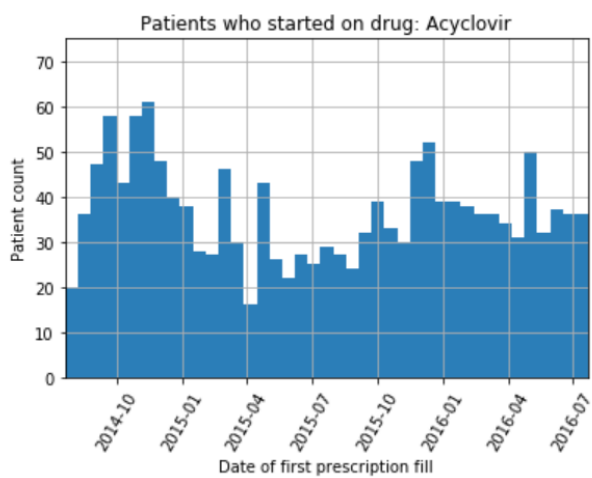

*Figure 4: Distribution of patients who started Acyclovir (in first 24 months of its approval)*

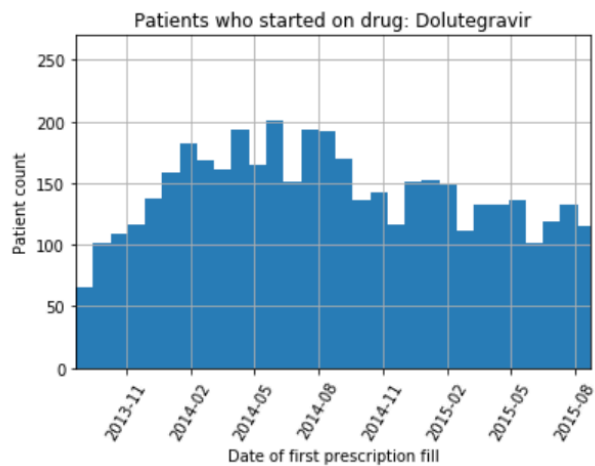

*Figure 5: Distribution of patients who started Dolutegravir (in first 24 months of its approval)*

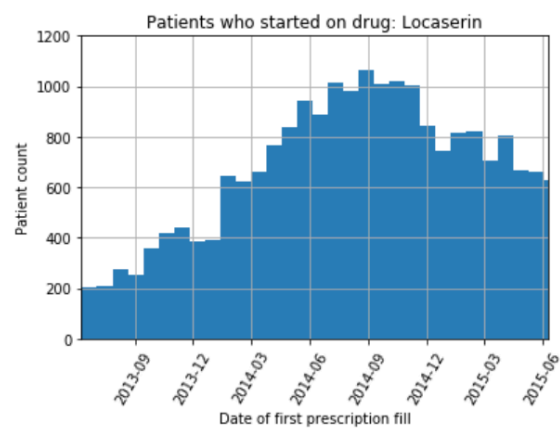

*Figure 6: Distribution of patients who started Locaserin (in first 24 months of its approval)*

## 4. DISCUSSION

In our study, we observed a significant difference in the mean number of observation days for those prescribed newly approved Drug A versus the standard of treatment Drug B. The average number of person-days on Drug A vs. Drug B was skewed and led to discovering a prescriber bias in treatment assignment. We reasoned that since the drug represented a new class of therapy not previously experienced by the doctors there was an adopter effect. Martin, et al discuss change in drug prescription pattern between clinical trials and post marketing. [8] Menzel, et al studied how social relations in a community of doctors can influence adoption of new treatments. [9] In addition, we believe the treatment was more likely in the beginning to be assigned to more severe patients. Stern, et al suggest that adoption of a new prescription drug also depends on prescription pattern of the prior drug category [10]. Many have [11,12] discussed evaluation of drugs in the real world and the potential biases. It is important that bias be assessed in the data before machine learning techniques are applied so that erroneous conclusions are not made. We tested multiple methods to control for the observed secular trend bias in our study, but residual bias remained such that machine learning could not be applied in our analysis.

As medical data grows in scope and depth, machine learning will be able to mine much needed insight from the data to positively impact clinical practice and patient health. However, data scientists need to be aware that potential heretofore unknown bias may lurk in the data. This bias needs to be identified and addressed in machine learning to minimize the uncertainty of AI. Being able to name and address the bias will also strengthen practitioners' adoption of the new knowledge approach of AI.

Our data show that prescriber patterns are influenced by many factors that when possible need to be accounted in models of treatment effect. We found that Drug A was not the only drug affected by secular trend bias in prescribing patterns. This could be due to higher cost of the newer drug, lack of physician experience with the drug especially when it represents a new drug class, or physician adoption readiness.